\documentclass[runningheads]{llncs}

\usepackage{eccv}

\usepackage{eccvabbrv}

\usepackage{graphicx}
\usepackage{booktabs}
\usepackage{amsmath}
\usepackage{amssymb}
 \usepackage{ulem}
\usepackage{mathtools}
\usepackage{multirow}
\usepackage{multicol}
\usepackage{lipsum}

\usepackage{colortbl}
\usepackage{enumitem}
\usepackage{caption}
\usepackage{wrapfig}
\usepackage{makecell}

\usepackage{pifont}
\newcommand{\cmark}{\ding{51}}%
\newcommand{\xmark}{\ding{55}}%
\definecolor{gainsboro}{RGB}{233,233,233}

\usepackage[accsupp]{axessibility}  
\newcommand{\proposed}{DPL} 

\usepackage{hyperref}

\usepackage{orcidlink}

\begin{document}

\title{Semantic Diversity-aware Prototype-based Learning for Unbiased Scene Graph Generation} 

\titlerunning{Semantic Diversity-aware Prototype-based Learning}

\author{Jaehyeong Jeon\orcidlink{0009-0005-7622-6991} \and
Kibum Kim\orcidlink{0000-0002-7381-019X} \and
Kanghoon Yoon\orcidlink{0000-0001-6947-2944} \and
Chanyoung Park\thanks{Corresponding author.}\orcidlink{0000-0002-5957-5816}
}

\authorrunning{J. Jeon et al.}

\institute{Korea Advanced Institute of Science and Technology (KAIST), South Korea \\ \email{\{wogud405, kb.kim, ykhoon08, cy.park\}@kaist.ac.kr}}

\maketitle

\begin{abstract}
  The scene graph generation (SGG) task involves detecting objects within an image and predicting predicates that represent the relationships between the objects. 
  However, in SGG benchmark datasets, each subject-object pair is annotated with a single predicate even though a single predicate may exhibit diverse semantics (i.e., semantic diversity), existing SGG models are trained to predict the one and only predicate for each pair. This in turn results in the SGG models to overlook the semantic diversity that may exist in a predicate, thus leading to biased predictions. 
  In this paper, we propose a novel model-agnostic Semantic \textbf{D}iversity-aware \textbf{P}rototype-based \textbf{L}earning (\proposed) framework that enables unbiased predictions based on the understanding of the semantic diversity of predicates. Specifically, DPL learns the regions in the semantic space covered by each predicate to distinguish among the various different semantics that a single predicate can represent. Extensive experiments demonstrate that our proposed model-agnostic DPL framework brings significant performance improvement on existing SGG models, and also effectively understands the semantic diversity of predicates. The code is available at \url{https://github.com/JeonJaeHyeong/DPL}.
\end{abstract}

\section{Introduction}
\label{sec:intro}
Scene Graph Generation (SGG) aims to predict relationships between objects within an image and generate a structured graph, in which the nodes and edges denote objects and pair-wise relationships between two objects, respectively. Since this graph contains high-level information of the an image, it is widely used in various downstream tasks such as image captioning~\cite{antol2015vqa, teney2017graph}, visual question answering~\cite{yang2019auto, gu2019unpaired}, and image retrieval~\cite{johnson2015image}. 

\begin{figure}[t]
\centering
    \centering
    \includegraphics[width=1.0\columnwidth]{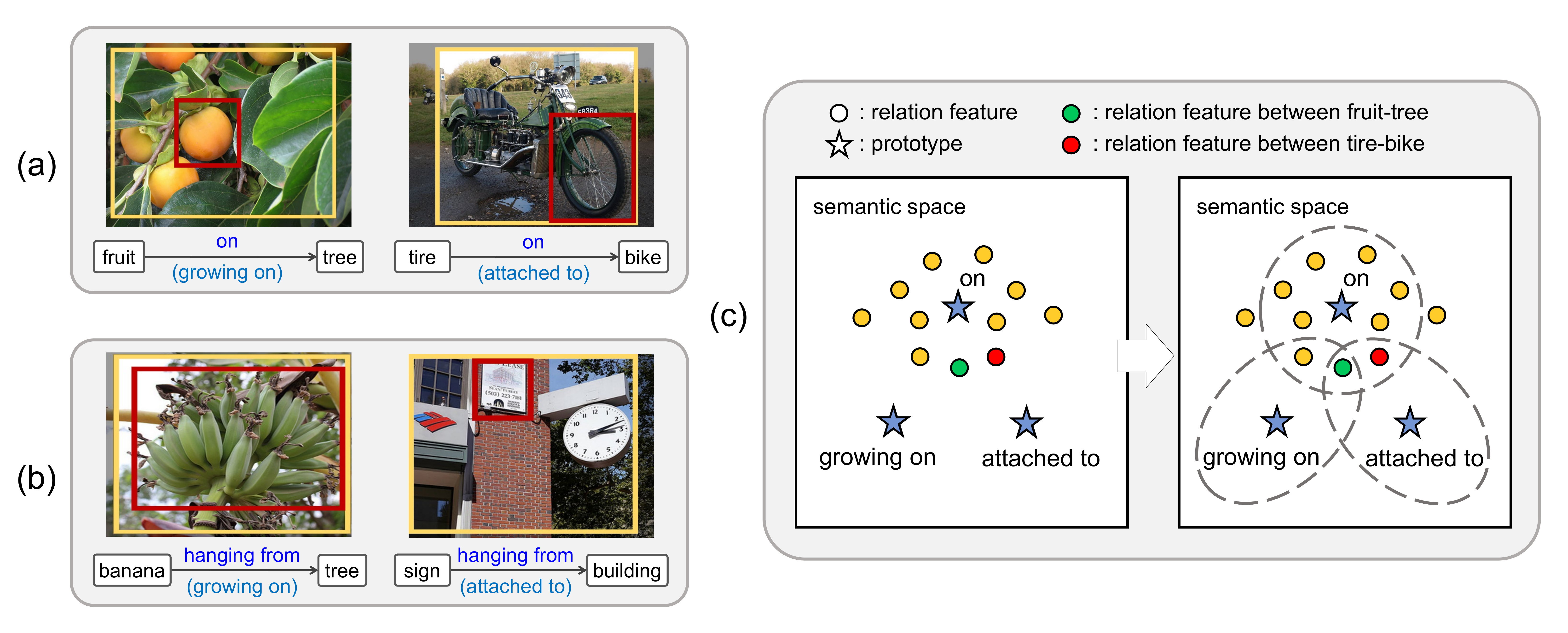}
    \captionof{figure}{Examples representing the semantic diversity of \textbf{(a)} \texttt{on}, and \textbf{(b)} \texttt{hanging from}. \textbf{(c)} Left: Due to the long-tail distribution of the dataset, many relation features are located near the prototype of the head class \texttt{on}. Right: The result of learning the regions representable by each predicate after our proposed DPL is adopted, where different semantics near the prototype \texttt{on} can be distinguished.}
\label{fig:intro}
\end{figure}

In this study, we focus on an inherent limitation of existing benchmark datasets for the SGG task (e.g., Visual Genome (VG)~\cite{krishna2017visual}). 
That is, each subject-object pair in an image is annotated with only a single predicate, even though a single predicate may exhibit diverse semantics, which we refer to as \textbf{semantic diversity}. 
For example, a predicate \texttt{on} generally denotes a subject being positioned \textit{on top of} an object. 
However, in the context of \texttt{<fruit, on, tree>}, \texttt{on} is used to express that the fruit is \textit{growing on} the tree, while in the context of \texttt{<tire, on, bike>}, \texttt{on} denotes that the  \texttt{tire} is \textit{attached to} the \texttt{bike} (See Fig.~\ref{fig:intro}(a)). It is important to note that such semantic diversity is not only prevalent in general predicates, such as \texttt{on}, but also in fine-grained predicates. 
{For example, as shown in Fig.~\ref{fig:intro}(b), the relationship between \texttt{banana-tree} and \texttt{sign-building} can both be described as \texttt{hanging from}. However, the relationship between \texttt{banana-tree} is associated with \textit{growing on}, while the relationship between \texttt{sign-building} is not related to \textit{growing on} and is rather closer to \textit{attached to}.}

Due to this inherent nature of the benchmark datasets, existing SGG models are trained to predict the one and only predicate annotated between each subject-object pair, and thus the semantic diversity that may exist in a predicate is overlooked. 
For example, despite the distant semantics of \texttt{on} in the two triplets \texttt{<fruit, on, tree>} and \texttt{<tire, on, bike>}, existing SGG models are trained to always predict \texttt{on}, thus leading to biased predictions. 
Although various unbiased SGG approaches have utilized re-balancing strategies such as re-sampling~\cite{li2021bipartite} or re-weighting~\cite{li2022ppdl, yan2020pcpl} to address the biased prediction issue, they mainly focus on training with instances in the tail classes, which eventually incurs a decrease in the head class performance or overfitting to the tail classes. Furthermore, recent research~\cite{li2022devil, zhang2022fine} has explored the re-labeling approach for converting the head classes into tail classes, but this approach results in the loss of the head class information. 
However, we argue that correctly capturing the semantic diversity within a predicate not only enriches the understanding of an image but also alleviates the problem caused by the long-tail predicate class distribution without having to explicitly manipulate between head and tail classes.

In this paper, we propose a novel model-agnostic Semantic \textbf{D}iversity-aware \textbf{P}rototype-based \textbf{L}earning (\proposed) framework that enables unbiased predictions based on the understanding of the semantic diversity of predicates. 
The main idea is to learn the regions in the semantic space covered by each predicate to distinguish among the various different semantics that a single predicate can represent (See Fig.~\ref{fig:intro}(c) right). More precisely, 
we introduce a learnable prototype for each predicate. For a given subject-object pair, we minimize the distance between its relation feature and the prototype corresponding to its ground-truth predicate.
As a result, we expect each prototype to be surrounded by relation features (dots in Fig.~\ref{fig:intro}(c)), whose relationship can be labeled with the predicate associated with the prototype. However, as the dataset exhibits a long-tailed predicate class distribution, most of the relation features tend to be placed near the prototypes of head classes (See \texttt{on} in Fig.~\ref{fig:intro}(c)), while not all relation features around \texttt{on} represent the same semantics due to the semantic diversity. For example, some of the relation features (e.g., red dot in Fig.~\ref{fig:intro}(c)) may correspond to \texttt{attached to}, while others (e.g., green dot in Fig.~\ref{fig:intro}(c)) are related to \texttt{growing on}. Besides, some relation features may exist near the prototype \texttt{on} due to the inherently biased predictions towards the head class, even through they cannot be expressed by \texttt{on}.

{
To capture the diverse semantics that a single predicate may exhibit, we generate samples around each prototype and match them with the relation features labeled with the predicate of the corresponding prototype.
By doing so, we can learn the region expressed by each predicate, and identify a certain semantic of subject-object relationships that can be expressed with the same predicate. For example, if there is a sample generated from \texttt{growing on} located close to a certain region near the prototype \texttt{on}, we can assume that this region is likely associated with the semantic of \texttt{growing on} {(e.g., overlapping area between the regions of \texttt{on} and \texttt{growing on} in Fig.~\ref{fig:intro}(c))}. {Similarly, we can estimate regions that are adjacent to the prototype \texttt{on} but are associated with \texttt{attached to} {(e.g., overlapping area between the regions of \texttt{on} and \texttt{attached to} in Fig.~\ref{fig:intro}(c)).} Additionally, by learning the actual region of the predicate \texttt{on}, we can prevent all relation features around \texttt{on} from being predicted as \texttt{on} even if they are not related to \texttt{on} {(e.g., 
yellow dots outside the region of \texttt{on} in Fig.~\ref{fig:intro}(c)).
}

In summary, our main contributions can be summarized as follows:
\begin{itemize}
\item We point out that existing SGG studies overlook the semantic diversity of predicates and emphasize its importance for informative scene graph generation.
\item We propose a model-agnostic \proposed{}, which employs prototypes and probabilistic sampling methods to capture the semantic diversity
\item Through extensive experiments on VG and GQA datasets, we demonstrate the effectiveness of the proposed method, achieving state-of-the-art performance.
\end{itemize}

\section{Related work}

\subsection{Scene Graph Generation}
The objective of SGG is to construct a graph that is useful for scene understanding by identifying the relationships between objects within an image. Early methods~\cite{lu2016visual, dai2017detecting, zhang2017visual} have overlooked the visual context and treated objects and pairwise relationships independently. Subsequent SGG approaches propose models that utilize rich contextual information by employing sequential LSTMs~\cite{zellers2018neural, tang2019learning}, message passing~\cite{xu2017scene, yang2018graph, li2021bipartite, yoon2023unbiased}, or tree structure modelling~\cite{tang2019learning}. Furthermore, ~\cite{gu2019scene, zareian2020bridging, sudhakaran2023vision} have attempt to overcome the limitations of SGG datasets by leveraging external knowledge. Recently, PE-Net~\cite{zheng2023prototype} attempted to obtain compact representations of entities and predicates using prototypes. However, these methods still overlook the semantic diversity of predicates and struggle with biased prediction problems.

\subsection{Unbiased Scene Graph Generation}
To address the biased prediction problem, numerous unbiased SGG works have been conducted. Similarly to\cite{li2022devil}, existing unbiased SGG methods can be broadly categorized into two groups: 1) Re-balancing based method: This involves increasing the number of tail classes or giving them more weight using techniques such as data augmentation~\cite{li2023compositional}, re-sampling~\cite{li2021bipartite}, or re-weighting\cite{yan2020pcpl, yu2020cogtree}. Moreover, recent research has explored re-labeling~\cite{li2022devil, zhang2022fine}, which involves transforming head classes into tail classes. These approaches can also be considered as re-balancing methods, as they involve adjusting the number of head and tail classes. 2) Unbiased prediction based on biased training: These models perform biased training but make unbiased predictions during inference. TDE~\cite{tang2020unbiased} introduces a causal inference framework to eliminate the context bias during the inference process. DLFE~\cite{chiou2021recovering} utilizes positive-unlabeled learning to recover unbiased probabilities. Our method belongs to the second category as we train the model to understand semantic diversity during biased training and apply this understanding during the inference phase. 

\section{Method}
Our goal is to learn the regions related to each predicate in the semantic space, enabling us to recognize the diverse semantics that a single predicate can express. To this end, we initially introduce prototypes to signify the representative semantics of predicates in the semantic space and train our model to minimize the distance between the relation feature and its ground truth class prototype (Section~\ref{subsec:prototype}). At the same time, we utilize a sampling approach to capture the diverse semantics represented by each predicate, contributing to our understanding of semantic diversity (Section~\ref{subsec:semantic_diversity}). Lastly, we make use of the semantic diversity information captured by our model to address the inevitable bias caused by the long-tailed predicate class distribution during the inference phase, leading to unbiased prediction (Section~\ref{subsec:unbiased_prediction}). 

\subsection{Preliminary}
Given an image $\mathit{I}$, the objective of SGG is to generate a scene graph $\mathcal{G} = \{ \mathit{O}, \mathit{E} \}$, where $\mathit{O}$ and $\mathit{E}$ denote the set of objects and thier pairwise relationships, respectively. The conventional SGG is generally conducted based on the following pipeline.

\subsubsection{Proposal Generation.} All objects $\mathit{O} =\{ \mathit{o}_i \}_{i=1}^{N_o}$ are identified in the image $\mathit{I}$ using a pre-trained object detector (e.g., Faster R-CNN~\cite{ren2015faster}), where $N_o$ is the number of objects. Each object $\mathit{o}_i \in \mathit{O}$ consists of a visual feature $\mathit{v}_i$, an object bounding box $\mathit{b}_i$, and an initial object label $\mathit{l}_i$. 
\subsubsection{Object Class Prediction.} Object features \(x_i\) are extracted based on the output of a proposal (i.e., $v_i$, $b_i$ and $w_i$), and object classification is conducted.
\begin{equation}
    x_i = f_{obj}(v_i, b_i, w_i),\,\,\,\,\hat{l}_i = \phi_{obj}(x_i),
\end{equation}
where \(w_i\) is the word embedding~\cite{pennington2014glove} of \(l_i\), \(f_{obj}\) is an object encoder such as BiLSTM~\cite{zellers2018neural} or fully connected layers~\cite{zhang2017visual}, and \(\phi_{obj}\) is an object classifier. An updated object label $\hat{l}_i$ is obtained in this procedure.
\subsubsection{Predicate Class Prediction.} This step involves obtaining a refined object feature \(\hat{x}_i\) using the word embedding \(\hat{w}_i\) of \(\hat{l}_i\). 
Then, given objects $o_i$ and $o_j$, and their features $\hat{x_i}$ and $\hat{x_j}$ along with their union feature \(u_{ij}\), we obtain its relation feature $r_{i\to j}$ that for predicate classification.
\begin{equation}
    \hat{x}_i = f_{rel}(v_i, x_i, \hat{w}_i), \,\,\,\,\mathit{r}_{i\to j} = f_{p}(\left[\hat{x_i}, \hat{x_j}\right]) \odot u_{ij}, \,\,\,\,\mathit{p}_{i\to j} = \phi_{rel}(\mathit{r}_{i\to j}),
\end{equation}
where \(\odot\) denotes element-wise product, \(f_{rel}\) is a relation encoder similar to \(f_{obj}\), \(f_{p}\) is fully-connected layer and \(\phi_{rel}\) is a predicate classifier. Moreover, $\mathit{p}_{i\to j}$ is the predicted predicate between objects $o_i$ and $o_j$, which is selected from a predefined set of predicates $\mathcal{P}$.

\begin{figure}[t]
\centering
    \centering
    \includegraphics[width=1.0\columnwidth]{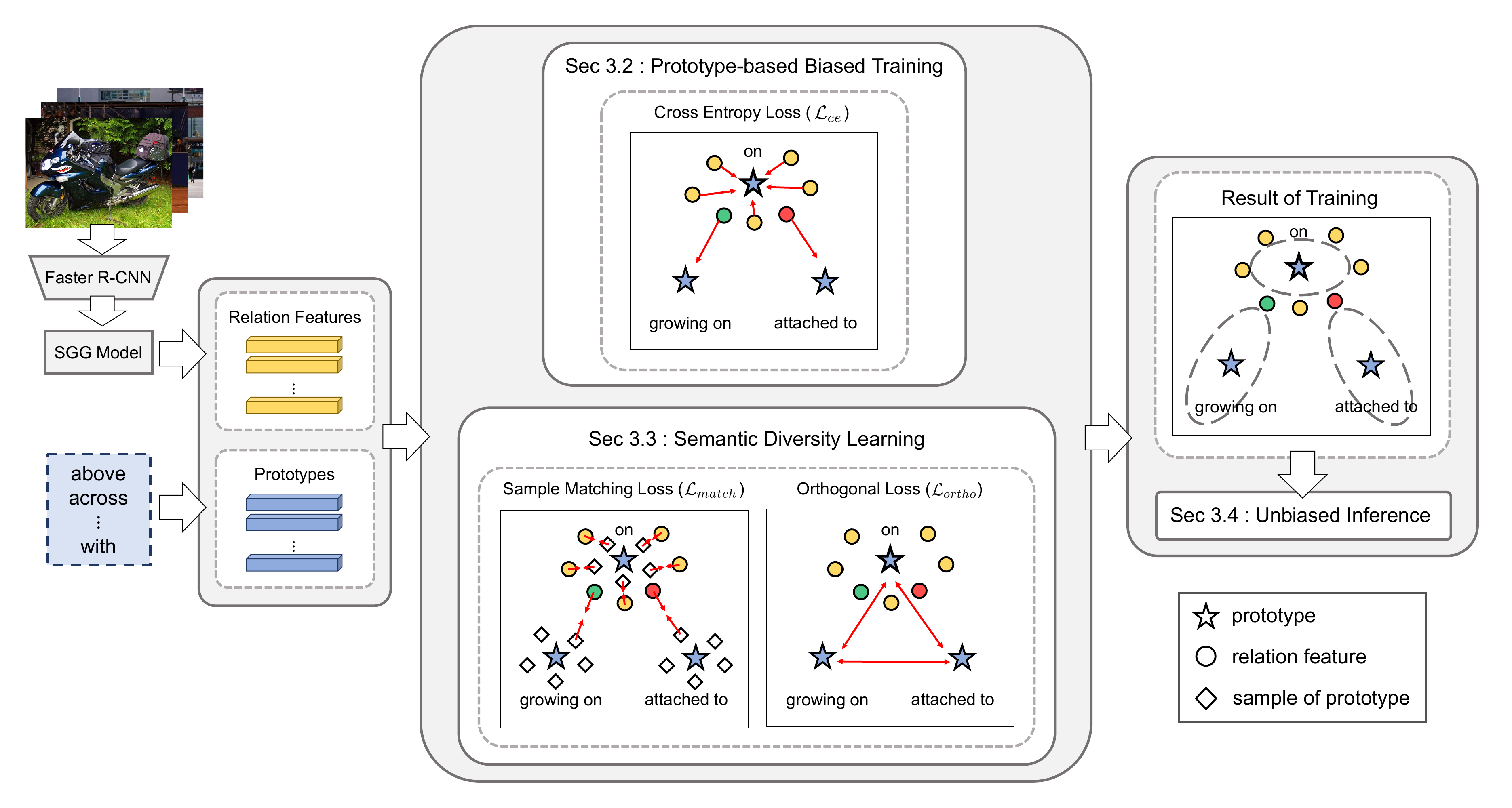}
    \captionof{figure}{The overall pipeline of Semantic Diversity aware Prototype-based Learning (\proposed{}) framework. First, \proposed{} obtains the relation feature from each subject-object pair, and creates a prototype corresponding to each predicate. Then, \proposed{} conducts prototype-based biased training to encourage each relation feature to approach its corresponding prototype (Sec~\ref{subsec:prototype}). A yellow dot corresponds to a relation feature labeled as \texttt{on}, while the green and red dots correspond to \texttt{growing on} and \texttt{attached to}, respectively.
    At the same time, \proposed{} conducts semantic diversity learning to capture regions that can be expressed by predicates (Sec~\ref{subsec:semantic_diversity}).
    These regions can be understood as the variances of prototypes. Lastly, we compute the normalized distance to perform unbiased inference (Sec.~\ref{subsec:unbiased_prediction}).
    }
\label{fig:model}
\end{figure}

\subsection{Prototype-based Biased Training}
\label{subsec:prototype} 
To handle representative semantics corresponding to each predicate, we first create prototypes $\textit{C} = \left\{c_1,c_2,\ldots,c_{|\mathcal{P}|} \right\}$, where \(c_i \in \mathbb{R}^{d}\) is the learnable prototype for the $i$-th predicate.
Furthermore, we use a trainable projector \(\phi_{proj} (\cdot)\) to align the relation feature $r$ with the same dimensional space as the prototype, i.e., \(z = \phi_{proj}(r)\), where \(z \in \mathbb{R}^{d}\).
The prototypes are $L_2$ normalized to ensure that they reside within a consistent range, i.e., \(\left\| c_i\right\|_2 = 1\). In this section, we aim to train each relation feature $z$ to become closer to the prototype of its ground-truth class. Specifically, we compute the probability of the relation represented by $z$ belonging to the $i$-th predicate class based on the Euclidean distance~\cite{chun2021probabilistic}:
\begin{equation}
    \textit{p}(i\text{-th class} \mid z) = \mathrm{Softmax}(-a\left\|z - c_i\right\|_2+b),
\end{equation}
where $a$ and $b$ are learnable scalars. Then, we train our model with the cross-entropy loss as follows:
\begin{equation}
     \mathcal{L}_{ce} = - \sum_{i=1}^{|\mathcal{P}|} y_{i} \log\textit{p}(i\text{-th class} \mid z), 
\end{equation}
where $y\in\{0,1\}^{|\mathcal{P}|}$ is a one-hot label vector of the relation feature $z$, where the element corresponding to the ground-truth predicate is 1.

Training with the above loss makes the relation feature $z$ close to its corresponding prototype. However, due to the long-tailed distribution of the predicate class, the relation features would be biased towards head classes, even though their ground truth predicates may not belong to the head classes (See Fig.~\ref{fig:model})

\subsection{Semantic Diversity Learning}
\label{subsec:semantic_diversity}

Recall that a single predicate may exhibit diverse semantics of subject-object relationships.
Therefore, it is necessary to understand the range that each predicate can represent and identify which parts within that range correspond to which semantics. 
For this purpose, we generate samples from each prototype to capture this phenomenon.
Specifically, we generate $N$ samples \(s_i = \{s_i^{(1)}, s_i^{(2)}, \dots, s_i^{(N)} \}\) around a prototype $c_i$ based on a normal distribution with the prototype as the mean and a diagonal covariance matrix induced by the prototype as follows:
\begin{align}
    & \mathrm{p}(s_i \mid c_i) \sim \mathcal{N}(\mu_i,\,\sigma_i^{2}) ,
\end{align}
where \(\mu_i = c_i\), \(\sigma_i^{2} = f_\sigma(c_i)\), and \(f_{\sigma}\) is a fully connected network.

\subsubsection{Sample Matching Loss.} 
To ensure that the samples fully cover the regions represented by each predicate, we introduce a matching function that connects each sample with its corresponding relation feature. Specifically, the matching loss is defined as:
\begin{align}
    & \mathcal{L}_{match} = \left( \max(0, \min_j\left\|z - s_k^{(j)}\right\|_2 - R) \right)^2,
\end{align}
where $R$ is a hyperparameter, and $k$ is the index of the ground-truth predicate class of the relation feature $z$. 
This enforces at least one of the samples of a given predicate to reside within the distance $R$ from the relation feature $z$ that is labeled with the given predicate. 

We would like to emphasize the importance of the hyperparameters $N$ and $R$. As shown in \cref{fig:model}, most of the relation features are located around the prototype \texttt{on}. From the perspective of the prototype \texttt{on}, relation features labeled as \texttt{on} are distributed all around. Therefore, it requires enough samples to learn regions in all directions (i.e., large $N$). However, from the perspective of the prototype \texttt{growing on}, the relation features labeled with \texttt{growing on} are not spread out but rather located around a specific region. Therefore, we can expect that capturing the region of \texttt{growing on} would be sufficient with relatively fewer samples (i.e., small $N$). Moreover, if $R$ is either too small or too large, the model may fail to learn the generalized regions that each predicate can represent. Detailed analysis demonstrating our intuition is shown in Section~\ref{sec:ablation}.

By leveraging the relationship between relation features and samples, we can differentiate between various semantics represented by the same predicate. 
For instance, if a relation feature near the prototype \texttt{on} closely aligns with the sample generated from \texttt{attached to}, then that feature would likely exhibit the semantic of \texttt{attached to}. Similarly, if a sample generated from \texttt{growing on} closely aligns with a particular relation feature, then that feature may be associated with the semantic of \texttt{growing on}. Therefore, through our sample-relation feature matching process, we not only learn the range that can be represented by each predicate, but also estimate the specific semantics associated with relation features.

\begin{figure}[t]
\centering
    \centering
    \includegraphics[width=0.9\columnwidth]{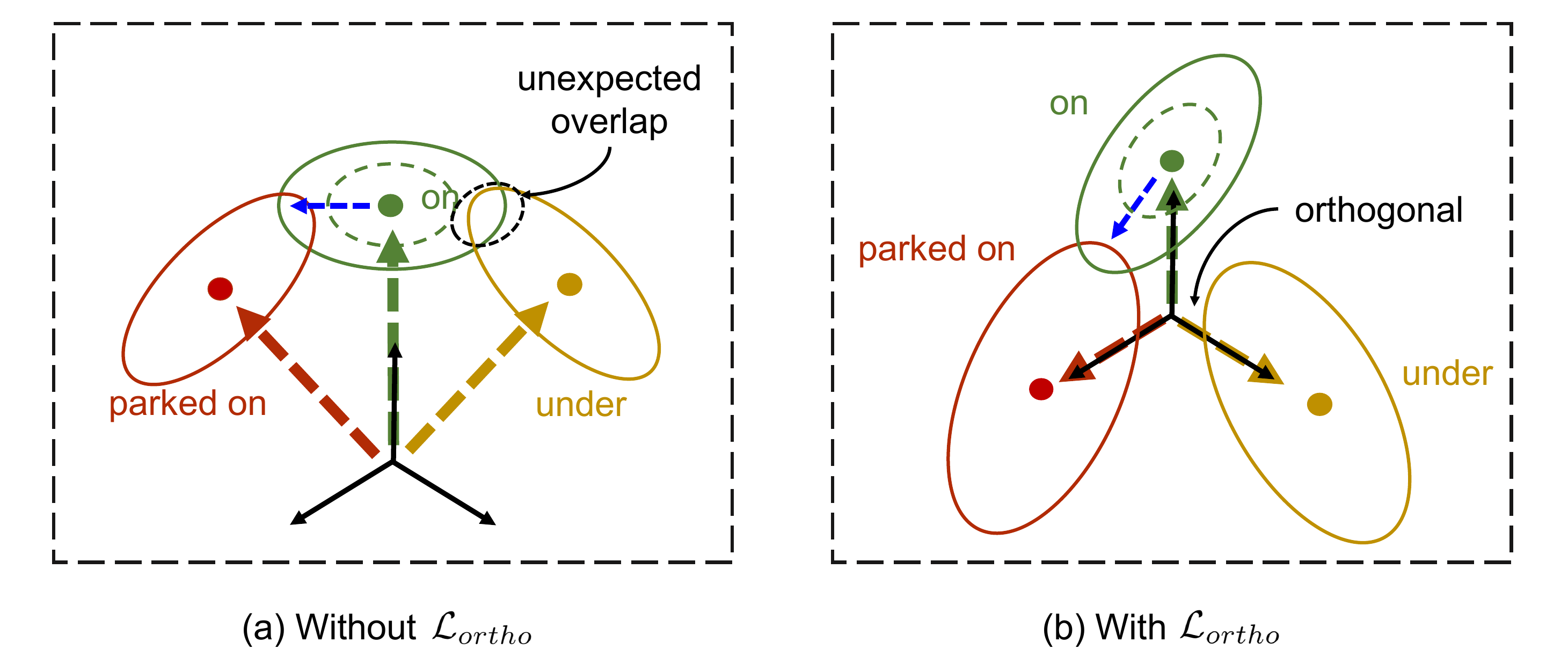}
    \captionof{figure}{Conceptual illustration of applying the orthogonal loss.
    \textbf{(a)} Without the orthogonal loss, there is a risk of an unexpected overlap between unrelated predicates due to the symmetry of the normal distribution. \textbf{(b)} However, applying the orthogonal loss can prevent such phenomena.}
\label{fig:ortho}
\end{figure}

\subsubsection{Orthogonal Loss.} Note that the distribution around each prototype shows a symmetric shape, as we have assumed it follows a normal distribution. As a result, if the variance increases in one direction, it can cause an unexpected overlap with other sample distributions as it also increases in the opposite direction. In Fig.~\ref{fig:ortho}, the variance of \texttt{on} extends towards \texttt{parked on} because \texttt{on} can express the semantic of \texttt{parked on}. However, the variance also increases in the opposite direction due to the symmetry of the normal distribution. This leads to an overlap with \texttt{under}, which cannot be used together with \texttt{on}. Therefore, to prevent such unexpected overlaps, an orthogonal loss is employed to make the prototypes independent from each other.
\begin{align}
    & \mathcal{L}_{ortho} = \frac{1}{|\mathcal{P}|(|\mathcal{P}|-1)} \sum_{i \neq j} \left| c_i \cdot c_j^T \right|,
\end{align}

The final training loss is defined as follows:
\begin{align}
    & \mathcal{L} = \mathcal{L}_{ce} + \mathcal{L}_{ortho} + \alpha \mathcal{L}_{match},
\end{align}
where $\alpha$ is hyperparameter.

\subsection{Unbiased Inference using Normalization}
\label{subsec:unbiased_prediction}
During training, the prediction of predicate classes relies on the Euclidean distance between the relation feature $z$ and its corresponding prototype $c_i$. {However, relying solely on Euclidean distance leads to biased predictions and overlooks semantic diversity of predicates. Therefore, during inference, we compute the normalized distance using the semantic diversity information encapsulated within $\sigma_i = \left[\sigma_i^{(1)}, \sigma_i^{(2)}, \dots , \sigma_i^{(d)} \right]$ as follows:}
{
\begin{equation}
    \textit{p}(i\text{-th class} \mid z) = \mathrm{Softmax}(-a' \left\| (z - c_j) \odot \sigma_j^{-1} \right\|_2+b), 
\end{equation}
where $\sigma_i^{-1} = \left[ \frac{1}{\sigma_i^{(1)}}, \frac{1}{\sigma_i^{(2)}}, \dots, \frac{1}{\sigma_i^{(d)}} \right]$ and $a' = a \cdot \frac{\max_j\left\| z - c_j \right\|_2}{\max_j\left\| (z - c_j) \odot \sigma_j^{-1} \right\|_2} $ mitigates the scaling effect caused by $\sigma_i$.
}
\section{Experiment}

\subsection{Experimental Settings}

\subsubsection{Datasets.} Experiments are conducted on two datasets: VG~\cite{krishna2017visual} and GQA~\cite{hudson2019gqa}. VG contains 108k images across 75k object categories and 37k predicate categories. Following previous studies on SGG~\cite{chen2019knowledge, chiou2021recovering, lin2020gps, li2021bipartite, suhail2021energy}, we adopt the most widely-used VG150 split~\cite{xu2017scene}, which contains the most frequent 150 object categories and 50 predicate categories. GQA is another vision-language dataset that contains over 3.8 million relation annotations. In the case of GQA, we follow the GQA200 split~\cite{dong2022stacked}, which select the 200 most frequent object categories and 100 most frequent predicate categories. For both datasets, we allocate 70\% of the data to the training set and 30\% to the testing set, and sample 5K images from the training set for validation.

\subsubsection{Tasks.} Following prior studies~\cite{xu2017scene, zellers2018neural}, we evaluate models on three conventional SGG tasks: 1) \textit{Predicate Classification} (\textbf{PredCls}): Predicting predicate class based on ground-truth object bounding boxes and their object classes. This task is not affected by the performance of the object detector. 2) \textit{Scene Graph Classification} (\textbf{SGCls}): Predicting both predicate classes and object classes, given the ground-truth bounding boxes. 3) \textit{Scene Graph Generation} (\textbf{SGGen}): Detecting object bounding boxes and predicting both the object classes and pairwise predicate classes. The bounding box is considered valid one if the IoU is over 0.5.

\subsubsection{Metrics.} For evaluation, we use Recall@K (R@K) and mean Recall@K (mR@K). Furthermore, following prior studies~\cite{zhang2022fine, min2023environment, kim2024adaptive,kim2024llm4sgg}, we adopt the harmonic average of R@K and mR@K, denoted as F@K. Generally, mR@K tends to have lower values in comparison to R@K. This indicates that a simple average between them would be heavily influenced by R@K. Instead, F@K provides a relatively fair comparison between R@K and mR@K as it gives more weight to smaller values.

\subsubsection{Implementation Details.} Following prior studies~\cite{tang2020unbiased}, we employ a pre-trained Faster R-CNN~\cite{ren2015faster} with ResNeXt-101-FPN~\cite{xie2017aggregated}. For the VG dataset, the pretrained object detector provided by~\cite{tang2020unbiased} is employed. 
{For the GQA dataset, due to the absence of an officially available object detector, we pretrain a new object detector. Therefore, we cannot directly compare our performance with other studies~\cite{dong2022stacked, sudhakaran2023vision} on the GQA dataset. The performance of the object detector used on the GQA dataset is 25.33 mAP. The initial learning rate is 0.01 and we use the SGD optimizer. The batch size is set to 3 on 1 GPU. For each of the three tasks, the training phase consists of 60,000 steps in total. Most of the hyperparameters align with those used in prior studies. The hyperparameter $\alpha$ is set to 10, and the best performance is achieved when $N$ and $R$ are set to 20 and 1.0, respectively. The dimension size of the relation feature and prototype is set to 128 (i.e., $d$=128).}

\subsection{Comparison with State-of-the-Arts}

\begin{table}[t]
\caption{Performance (\%) comparisons on the VG dataset. +\textbf{DPL} denotes different models equipped with our proposed DPL.}

\centering
\resizebox{1.0\textwidth}{!}{
\setlength{\tabcolsep}{3.5pt}
\renewcommand{\arraystretch}{1.25}
\begin{tabular}{l|ccc|ccc|ccc}
    \toprule
    \multicolumn{1}{c|}{\multirow{2}{*}{Model}} & \multicolumn{3}{c|}{PredCls}                                                                 & \multicolumn{3}{c|}{SGCls}                                                                   & \multicolumn{3}{c}{SGDet}                                                                  \\ \cmidrule{2-10} 
    \multicolumn{1}{c|}{}                       & \multicolumn{1}{l}{R@50 / 100} & \multicolumn{1}{l}{mR@50 / 100} & \multicolumn{1}{l|}{F@50 / 100} & \multicolumn{1}{l}{R@50 / 100} & \multicolumn{1}{l}{mR@50 / 100} & \multicolumn{1}{l|}{F@50 / 100} & \multicolumn{1}{l}{R@50 / 100} & \multicolumn{1}{l}{mR@50 / 100} & \multicolumn{1}{l}{F@50 / 100} \\ \midrule
    IMP~\cite{xu2017scene}                                         & 61.1 / 63.1                  & 11.0 / 11.8                   & 18.6 / 19.9                   & 37.4 / 38.3                  & 6.4 / 6.7                     & 10.9 / 11.4                   & 23.6 / 28.7                  & 3.3 / 4.1                     & 5.8 / 7.2                    \\
    KERN~\cite{chen2019knowledge}                                        & 65.8 / 67.6                  & 17.7 / 19.2                   & 27.9 / 29.9                   & 36.7 / 37.4                  & 9.4 / 10.0                    & 15.0 / 15.8                   & 27.1 / 29.8                  & 6.4 / 7.3                     & 10.4 / 11.7                  \\
    GPS-Net~\cite{lin2020gps}                                     & 65.2 / 67.1                  & 15.2 / 16.6                   & 24.7 / 26.6                   & 37.8 / 39.2                  & 8.5 / 9.1                     & 13.9 / 14.8                   & 31.1 / 35.9                  & 6.7 / 8.6                     & 11.0 / 13.9                  \\
    BGNN~\cite{li2021bipartite}                                        & 59.2 / 61.3                  & 30.4 / 32.9                   & 40.2 / 42.8                   & 37.4 / 38.5                  & 14.3 / 16.5                   & 20.7 / 23.1                   & 31.0 / 35.8                  & 10.7 / 12.6                   & 15.9 / 18.6                  \\     SQUAT~\cite{jung2023devil}                                        & 55.7 / 57.9                  & 30.9 / 33.4                   & 39.7 / 42.4                   & 33.1 / 34.4                  & 17.5 / 18.8                   & 22.9 / 24.3                   & 24.5 / 28.9                  & 14.1 / 16.5                   & 17.9 / 21.0                  \\ \midrule
    VTransE~\cite{zhang2017visual}                                     & \textbf{65.7} / \textbf{67.6}         & 14.7 / 15.8                   & 24.0 / 25.6                   & \textbf{38.6} / \textbf{39.4}         & 8.2 / 8.7                     & 13.5 / 14.3                   & \textbf{29.7} / \textbf{34.3}         & 5.0 / 6.0                     & 8.6 / 10.2                   \\
    + TDE~\cite{tang2020unbiased}                                       & 43.1 / 48.5                  & 24.6 / 28.0                   & 31.3 / 35.5                   & 25.7 / 28.5                  & 12.9 / 14.8                   & 17.2 / 19.5                   & 18.7 / 22.6                  & 8.6 / 10.5                    & 11.8 / 14.3                  \\
    \rowcolor{gainsboro} + \textbf{\proposed}                                       & 56.2 / 58.0                  & \textbf{33.3} / \textbf{36.3}           & \textbf{41.8} / \textbf{44.7}          & 30.4 / 32.9                  & \textbf{16.3} / \textbf{18.2}          & \textbf{21.2} / \textbf{23.4}          & 19.4 / 24.2                  & \textbf{11.2} / \textbf{13.7}          & \textbf{14.2} / \textbf{17.5}         \\ \midrule
    Motifs~\cite{zellers2018neural}                                      & \textbf{65.2} / \textbf{67.0}         & 14.8 / 16.1                   & 24.1 / 26.0                   & \textbf{38.9} / \textbf{39.8}         & 8.3 / 8.8                     & 13.7 / 14.8                   & \textbf{32.8} / \textbf{37.2}         & 6.8 / 7.9                     & 11.0 / 13.9                  \\
    + Rwt~\cite{chiou2021recovering}                                      & 54.7 / 56.5                  & 17.3 / 18.6                   & 26.3 / 28.0                   & 29.5 / 31.5                  & 11.2 / 11.7                     & 16.2 / 17.1                   & 24.4 / 29.3                  & 9.2 / 10.9                     & 13.4 / 15.9                  \\
    
    + TDE~\cite{tang2020unbiased}                                       & 46.2 / 51.4                  & 25.5 / 29.1                   & 32.9 / 37.2                   & 27.7 / 29.9                  & 13.1 / 14.9                   & 17.8 / 19.9                   & 16.9 / 20.3                  & 8.2 / 9.8                     & 11.0 / 13.2                  \\
    + DLFE~\cite{chiou2021recovering}                                      & 52.5 / 54.2                  & 26.9 / 28.8                   & 35.6 / 37.6                   & 32.3 / 33.1                  & 15.2 / 15.9                   & 20.7 / 21.5                   & 25.4 / 29.4                  & 11.7 / 13.8                   & 16.0 / 18.8                  \\
    + CogTree~\cite{yu2020cogtree}                                   & 35.6 / 36.8                  & 26.4 / 29.0                   & 30.3 / 32.4                   & 21.6 / 22.2                  & 14.9 / 16.1                   & 17.6 / 18.7                   & 20.0 / 22.1                  & 10.4 / 11.8                   & 13.7 / 15.4                  \\
    + PCPL~\cite{yan2020pcpl}                                      & 54.7 / 56.5                  & 24.3 / 26.1                   & 33.7 / 35.7                   & 35.3 / 36.1                  & 12.0 / 12.7                   & 17.9 / 18.8                   & 27.8 / 31.7                  & 10.7 / 12.6                   & 15.5 / 18.0                  \\
    + IETrans~\cite{zhang2022fine}                                   & 54.7 / 56.7                  & 30.9 / 33.6                   & 39.5 / 42.2                   & 32.5 / 33.4                  & 16.8 / 17.9                   & 22.2 / 23.3                   & 26.4 / 30.6                  & 12.4 / 14.9                   & 16.9 / 20.0                  \\
    \rowcolor{gainsboro} + \textbf{\proposed}                                        & 54.4 / 56.3                  & \textbf{33.7 / 37.4}          & \textbf{41.6 / 44.9}          & 32.6 / 33.8                  & \textbf{18.5 / 20.1}          & \textbf{23.6 / 25.2}          & 24.5 / 28.7                  & \textbf{13.0 / 15.6}          & \textbf{17.0 / 20.2}         \\ \midrule
    VCTree~\cite{tang2019learning}                                      & \textbf{65.4} / \textbf{67.2}         & 16.7 / 18.2                   & 26.6 / 28.6                   & \textbf{46.7} / \textbf{47.6}         & 11.8 / 12.5                   & 18.8 / 19.8                   & \textbf{31.9} / \textbf{36.2}         & 7.4 / 8.7                     & 12.0 / 14.0                  \\
    + Rwt~\cite{chiou2021recovering}                                
        & 60.7 / 62.6         & 19.4 / 20.4                   
        & 29.4 / 30.8                   & 42.3 / 43.5         
        & 12.5 / 13.1                   & 19.3 / 20.1                   & 27.8 / 32.0         & 8.7 / 10.1                     
        & 13.3 / 15.4                  \\
    + TDE~\cite{tang2020unbiased}                                       & 47.2 / 51.6                  & 25.4 / 28.7                   & 33.0 / 36.9                   & 25.4 / 27.9                  & 12.2 / 14.0                   & 16.5 / 18.6                   & 19.4 / 23.2                  & 9.3 / 11.1                    & 12.6 / 15.1                  \\
    + DLFE~\cite{chiou2021recovering}                                      & 51.8 / 53.5                  & 25.3 / 27.1                   & 34.0 / 36.0                   & 33.5 / 34.6                  & 18.9 / 20.0                   & \textbf{24.2} / 25.3                   & 22.7 / 26.3                  & 11.8 / 13.8                   & 15.5 / 18.1                  \\
    + CogTree~\cite{yu2020cogtree}                                   & 44.0 / 45.4                  & 27.6 / 29.7                   & 33.9 / 35.9                   & 30.9 / 31.7                  & 18.8 / 19.9                   & 23.4 / 24.5                   & 18.2 / 20.4                  & 10.4 / 12.1                   & 13.2 / 15.2                  \\
    + PCPL~\cite{yan2020pcpl}                                      & 56.9 / 58.7                  & 22.8 / 24.5                   & 32.6 / 34.6                   & 40.6 / 41.7                  & 15.2 / 16.1                   & 22.1 / 23.2                   & 26.6 / 30.3                  & 10.8 / 12.6                   & 15.4 / 17.8                  \\
    + IETrans~\cite{zhang2022fine}                                   & 53.0 / 55.0                  & 30.3 / 33.9                   & 38.6 / 41.9                   & 32.9 / 33.8                  & 16.5 / 18.1                   & 22.0 / 23.6                   & 25.4 / 29.3                  & 11.5 / 14.0                   & 15.8 / 18.9                  \\
    \rowcolor{gainsboro} + \textbf{\proposed}                                       & 54.0 / 55.8                  & \textbf{35.3 / 37.9}          & \textbf{42.7 / 45.1}          & 32.6 / 33.9                  & \textbf{19.2 / 21.3}          & \textbf{24.2 / 26.2}          & 27.0 / 31.4                  & \textbf{13.2 / 15.9}          & \textbf{17.7 / 21.1}         \\  \bottomrule
    \end{tabular}
}

\label{tab:VG_table}
\end{table}

\begin{table*}[t]
\caption{Performance comparison of different methods on the GQA dataset. }
\centering
\resizebox{1.0\textwidth}{!}{
\setlength{\tabcolsep}{4pt}
\renewcommand{\arraystretch}{1.25}
\begin{tabular}{l|ccc|ccc|ccc}
    \toprule
    \multicolumn{1}{c|}{\multirow{2}{*}{Model}} & \multicolumn{3}{c|}{PredCls}                                                                 & \multicolumn{3}{c|}{SGCls}                                                                   & \multicolumn{3}{c}{SGDet}                                                                  \\ \cmidrule{2-10} 
    \multicolumn{1}{c|}{}                       & \multicolumn{1}{l}{R@50 / 100} & \multicolumn{1}{l}{mR@50 / 100} & \multicolumn{1}{l|}{F@50 / 100} & \multicolumn{1}{l}{R@50 / 100} & \multicolumn{1}{l}{mR@50 / 100} & \multicolumn{1}{l|}{F@50 / 100} & \multicolumn{1}{l}{R@50 / 100} & \multicolumn{1}{l}{mR@50 / 100} & \multicolumn{1}{l}{F@50 / 100} \\ \midrule
    VTransE~\cite{zhang2017visual}   & \textbf{65.0} / \textbf{66.5}    & 14.9 / 15.8        & 24.2 / 25.5                          & \textbf{30.2} / \textbf{31.1}    & 6.9 / 7.3          & 11.2 / 11.8                          & \textbf{26.7} / \textbf{30.6}    & 5.3 / 6.2          & 8.8 / 10.3                  \\
    \rowcolor{gainsboro} +\textbf{\proposed}    
    & 47.0 / 51.5     & \textbf{30.9} / \textbf{33.8}          & \textbf{37.3} / \textbf{40.8}                      & 24.1 / 25.2                  & \textbf{13.3} / \textbf{14.2}          & \textbf{17.1} / \textbf{18.2}                      & 21.1 / 24.8                  & \textbf{10.7} / \textbf{12.7}          & \textbf{14.2} / \textbf{16.8}         \\ \midrule
    Motifs~\cite{zellers2018neural}   & \textbf{64.9} / \textbf{66.4}         & 14.9 / 15.6                   & 24.2 / 25.3          & \textbf{30.1} / \textbf{30.9}         & 6.2 / 6.5                     & 10.3 / 10.7          & \textbf{22.7} / \textbf{27.3}         & 4.2 / 5.2                     & 7.1 / 8.7                    \\
    \rowcolor{gainsboro} + \textbf{\proposed}                          & 50.3 / 52.3    & \textbf{31.6} / \textbf{33.9}   & \textbf{38.8} / \textbf{41.1}   & 25.0 / 25.9    & \textbf{13.3} / \textbf{14.4}   & \textbf{17.4} / \textbf{18.5}   & 15.0 / 19.0    & \textbf{11.1} / \textbf{13.1}   & \textbf{12.8} / \textbf{15.5}         \\ \midrule
    VCTree~\cite{tang2019learning}    
    & \textbf{65.0} / \textbf{66.6}   & 15.7 / 16.6    & 25.3 / 26.6   
    & \textbf{27.8} / \textbf{28.9}   & 6.1 / 6.5      & 10.0 / 10.6   & \textbf{21.6} / \textbf{25.5}   & 4.1 / 5.1      & 6.9 / 8.5                    \\
    \rowcolor{gainsboro} + \textbf{\proposed}                                       & 46.1 / 50.6      & \textbf{31.4} / \textbf{34.3}    & \textbf{37.4} / \textbf{40.9}        & 22.1 / 23.7      & \textbf{13.3} / \textbf{14.9}    & \textbf{16.6} / \textbf{18.3}        & 14.4 / 18.1      & \textbf{11.9} / \textbf{13.7}    & \textbf{13.0 / 15.6}        \\  \bottomrule
\end{tabular}
}

\label{tab:GQA_table}
\end{table*}

\subsubsection{Baselines.} To demonstrate the effectiveness of our model-agnostic approach, we applied our method to three baseline SGG models: VTransE~\cite{zhang2017visual}, Motifs~\cite{zellers2018neural}, and VCTree~\cite{tang2019learning}. We compared our method with the state-of-the-art models, including IMP~\cite{xu2017scene}, KERN~\cite{chen2019knowledge}, GPS-Net~\cite{lin2020gps}, BGNN~\cite{li2021bipartite}, SQUAT~\cite{jung2023devil}. Additionally, we conducted comparisons with model-agnostic SGG methods, which can be categorized into two groups: 1) \textbf{Unbiased SGG from biased training}: TDE~\cite{tang2020unbiased} and DLFE~\cite{chiou2021recovering}. These models perform unbiased inference from biased training, utilizing counterfactual causality and frequency estimation, respectively. 2) \textbf{Re-balancing based method}: CogTree~\cite{yu2020cogtree}, PCPL~\cite{yan2020pcpl}, Re-weighting~\cite{chiou2021recovering}, and IETrans~\cite{zhang2022fine}. These methods employ re-balancing techniques such as a re-weighting-based approach or a re-labeling-based approach.

\subsubsection{Quantitative Results.} Table~\ref{tab:VG_table} shows the performance comparison with various SGG models on the VG dataset. We observe that \proposed{} greatly improves the performance of the baseline models, demonstrating the effectiveness of our model-agnostic approach. Furthermore, when compared to unbiased SGG methods, DPL outperforms not only TDE and DLFE, which perform unbiased prediction from biased training, but also recent re-balancing methods that have been extensively researched. This suggests that despite biased training, achieving outstanding performance is possible by understanding semantic diversity of predicates and effectively integrating them during inference phase. Additionally, DPL significantly improves the performance of the baseline models on the GQA dataset (Table~\ref{tab:GQA_table}), indicating that \proposed{} is a dataset-independent method.

\subsection{Ablation Studies}
\label{sec:ablation}

 We have two main hyperparameters, i.e., the number of samples $N$ and the value of $R$ in $L_{match}$. These hyperparameters are closely associated with the model's performance, and the related experimental results are presented in Table~\ref{tab:N_table} and Table~\ref{tab:R_table}. We also conducted experiments to evaluate the effect of each component in~\proposed, which are presented in Table~\ref{tab:comp_table}. These tables display the results of the PredCls task in the Motifs~\cite{zellers2018neural} baseline model on the VG dataset.
 
 \subsubsection{Influence of Hyperparameter} $N$. 
 First, an observation from Table~\ref{tab:N_table} is that as $N$ increases, the overall R@K value increases, while mR@K decreases. Especially, we can see a significant change in the performance of R@K, which mainly reflects the performance of head classes. This suggests that a large number of samples are required to capture the semantic diversity of the head classes. Conversely, even with $N$=1, the mR@K, which represents the performance of tail classes, is sufficiently high. This indicates that a smaller number of samples is enough to capture the semantic diversity of tail classes.
 Furthermore, due to the trade-off between R@K and mR@K, mR@K slightly decreases as $N$ increases. This decrease is not significant compared to the increase in R@K because increasing $N$ does not result in a loss of semantic diversity information for tail classes. Consequently, the overall F@K peaks when $N$ is round 20.
 
\begin{table}[t]
    \centering
    \begin{minipage}[t]{0.5\textwidth}
        \centering
        \setlength{\tabcolsep}{5pt} 
        \renewcommand{\arraystretch}{1.2}
        \caption{Ablation studies on $N$.}
        \resizebox{0.97\textwidth}{!}{
            \begin{tabular}{c|ccc}
                \toprule
                \multirow{2}{*}{$N$} & \multicolumn{3}{c}{PredCls}                                                                                 \\ \cline{2-4} 
                                   & \multicolumn{1}{l}{R@50/100}    & \multicolumn{1}{l|}{mR@50/100}            & \multicolumn{1}{l}{F@50/100} \\ \midrule
                1                  & \multicolumn{1}{c|}{40.6 / 42.4} & \multicolumn{1}{c|}{\textbf{36.4 / 40.1}} & 38.4 / 41.2                   \\
                5                  & \multicolumn{1}{c|}{50.5 / 52.6} & \multicolumn{1}{c|}{34.4 / 38.4}          & 40.9 / 44.4                   \\
                10                 & \multicolumn{1}{c|}{52.1 / 54.0} & \multicolumn{1}{c|}{34.1 / 37.7}          & 41.2 / 44.4                   \\
                20                 & \multicolumn{1}{c|}{54.4 / 56.3} & \multicolumn{1}{c|}{33.7 / 37.4}          & 41.6 / \textbf{44.9}          \\
                40                 & \multicolumn{1}{c|}{\textbf{56.8} / \textbf{58.6}} & \multicolumn{1}{c|}{33.0 / 36.0}          & \textbf{41.7} / 44.6                   \\ \bottomrule
            \end{tabular}
        }
        \label{tab:N_table}
    \end{minipage}%
    \hfill
    \begin{minipage}[t]{0.5\textwidth}
        \centering
        \setlength{\tabcolsep}{5pt} 
        \renewcommand{\arraystretch}{1.2}
        \caption{Ablation studies on $R$ in $L_{match}$.}
        \resizebox{\textwidth}{!}{
            \begin{tabular}{c|ccc}
            \toprule
            \multirow{2}{*}{$R$} & \multicolumn{3}{c}{PredCls}                                                                        \\ \cline{2-4} 
                               & \multicolumn{1}{l|}{R@50/100}    & \multicolumn{1}{l|}{mR@50/100}   & \multicolumn{1}{l}{F@50/100} \\ \midrule
            0.6                & \multicolumn{1}{c|}{42.6 / 45.5} & \multicolumn{1}{c|}{25.0 / 29.5} & 31.5 / 35.8                   \\
            0.8                & \multicolumn{1}{c|}{52.3 / 57.7} & \multicolumn{1}{c|}{24.0 / 28.3} & 32.9 / 38.0                   \\
            1.0                & \multicolumn{1}{c|}{54.4 / 56.3} & \multicolumn{1}{c|}{\textbf{33.7} / \textbf{37.4}} & \textbf{41.6} / \textbf{44.9}          \\
            1.2                & \multicolumn{1}{c|}{60.2 / 62.1} & \multicolumn{1}{c|}{27.6 / 31.3} & 37.8 / 41.6                   \\
            1.4                & \multicolumn{1}{c|}{\textbf{63.5} / \textbf{65.5}} & \multicolumn{1}{c|}{20.9 / 22.9} & 31.4 / 33.9                   \\ \bottomrule
            \end{tabular}
        }
        \label{tab:R_table}
    \end{minipage}
\end{table}
 
 \begin{wraptable}{r}[.001in]{0.5\textwidth}
    \centering
    \caption{Ablation studies of different $N$ assigned to each predicate.}
    \setlength{\tabcolsep}{5pt} 
    \renewcommand{\arraystretch}{1.2}
    \resizebox{0.5\textwidth}{!}{
        \begin{tabular}{c|ccc}
            \toprule
            \multirow{2}{*}{$N$} & \multicolumn{3}{c}{PredCls}                                                                                 \\ \cline{2-4} 
                               & \multicolumn{1}{l}{R@50/100}    & \multicolumn{1}{l|}{mR@50/100}            & \multicolumn{1}{l}{F@50/100} \\ \midrule
            Case 1 (Head > Tail)                  & \multicolumn{1}{c|}{\textbf{56.7} / \textbf{58.5}} & \multicolumn{1}{c|}{32.1 / 35.4} & \textbf{41.0} / \textbf{44.1}                   \\
            Case 2 (Head < Tail)                 & \multicolumn{1}{c|}{40.3 / 42.2} & \multicolumn{1}{c|}{\textbf{35.5} / \textbf{39.0}}          & 37.7 / 40.5                   \\ 
            \bottomrule        
        \end{tabular}
    }
    \label{tab:Nd_table}
\end{wraptable}
To gain a more precise understanding of the impact of $N$, we conduct experiments where different $N$ values are assigned to each predicate.
We compare Case 1, where head classes are assigned larger $N$ while tail classes receive smaller $N$, with Case 2, where tail classes are allocated larger $N$ and head classes receive smaller $N$. The total sum of $N$s of each predicate remains consistent across both cases, and the results are presented in Table~\ref{tab:Nd_table}. Although the total number of samples used is the same in both cases, there is a significant difference in the F@K between Case 1 and Case 2. This is because the head classes are more affected by $N$, resulting in a greater difference in R@K than in mR@K. Consequently, the overall F@K is much higher in Case 1. 
 
 \subsubsection{Influence of Hyperparameter} $R$. Table~\ref{tab:R_table} shows that performance varies depending on the value of $R$. When $R$ is set to a small or large value, a decrease in performance is observed. This phenomenon occurs because a small $R$ makes the model highly sensitive to even slight variations in relation features, potentially leading to overfitting. Conversely, when $R$ is large, most relation features come within a distance of $R$ from the samples, preventing the model from learning an appropriate variance.

\begin{table}[b]
    \caption{Ablation study on each component of DPL.}    
    \centering
    \setlength{\tabcolsep}{5pt} 
    \renewcommand{\arraystretch}{1.2}
    \resizebox{0.8\textwidth}{!}{
        \begin{tabular}{cccc|ccc}
        \toprule
        \multicolumn{4}{c|}{Components of DPL} & \multicolumn{3}{c}{PredCls}                                               \\ \midrule
        $\mathcal{L}_{ce}$   & $\mathcal{L}_{ortho}$  & $\mathcal{L}_{match}$  & \makecell{Unbiased \\ Inference}  & \multicolumn{1}{l|}{R@50/100} & \multicolumn{1}{l|}{mR@50/100} & \multicolumn{1}{l}{F@50/100} \\ \hline
        \cmark    & \xmark    & \xmark    & \xmark   & \multicolumn{1}{l|}{65.5 / 67.3}         & \multicolumn{1}{c|}{17.3 / 18.7}   & \multicolumn{1}{c}{27.4 / 29.3}         \\
        \cmark    & \cmark    & \xmark    & \xmark   & \multicolumn{1}{c|}{65.3 / 67.1}         & \multicolumn{1}{c|}{16.7 / 18.1}   & \multicolumn{1}{l}{26.6 / 28.5}         \\
        \cmark    & \xmark    & \cmark    & \xmark   & \multicolumn{1}{c|}{65.1 / 67.1}         & \multicolumn{1}{c|}{17.2 / 18.8}   & \multicolumn{1}{c}{27.2 / 29.4}         \\
        \cmark    & \cmark    & \cmark    & \xmark   & \multicolumn{1}{c|}{65.1 / 67.1}         & \multicolumn{1}{c|}{17.2 / 18.8}   & \multicolumn{1}{c}{27.2 / 29.4}         \\
        \cmark    & \xmark    & \cmark    & \cmark   & \multicolumn{1}{c|}{63.0 / 64.8}         & \multicolumn{1}{c|}{24.7 / 26.9}   & \multicolumn{1}{c}{26.9 / 35.5}         \\
        \cmark    & \cmark    & \cmark    & \cmark   & \multicolumn{1}{c|}{54.4 / 56.3}         & \multicolumn{1}{c|}{\textbf{33.7} / \textbf{37.4}}   & \multicolumn{1}{c}{\textbf{41.6} / \textbf{44.9}}         \\ \bottomrule
        \end{tabular}
    }
    \label{tab:comp_table}

\end{table}

\subsubsection{Effectiveness of each component.} Table~\ref{tab:comp_table} presents the results of comparing the influence of each component. A clear observation is that even when employing $L_{ortho}$ or $L_{match}$, their influence on performance improvement is minimal unless unbiased inference is employed. This signifies that even if the model learns a suitable variance of each predicate, the failure to incorporate it during inference prevents the ability to conduct unbiased SGG. The second observation is the significant impact of the orthogonal loss on learning an appropriate variance. Even when making unbiased inference without using the orthogonal loss, we observe a performance drop compared with when the orthogonal loss is adopted. This is because, as mentioned earlier, unexpected overlap can occur. Therefore, it is necessary to use the matching loss and the orthogonal loss together to effectively learn the appropriate variance.

\subsection{Qualitative Analysis}

\begin{figure*}[t]
\centering
    \centering
    \includegraphics[width=\columnwidth]{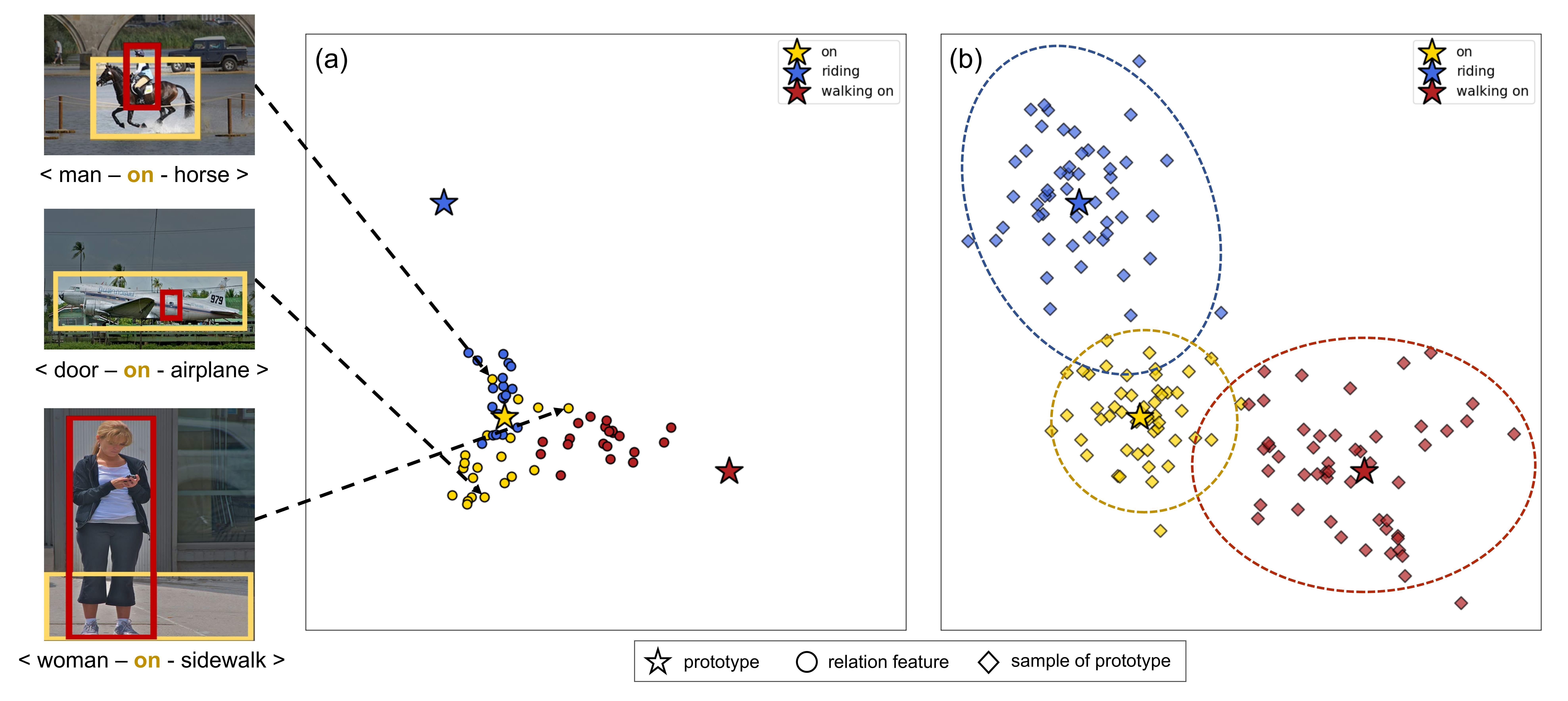}
    \captionof{figure}{The PCA visualization of \textbf{(a)} prototypes and relation features of subject-object pairs, and \textbf{(b)} prototypes and samples of each prototype. To better illustrate the trend of the learned variances, we equally scale up the variances to generate samples. The representations are obtained from DPL with Motifs backbone.
    }
\label{fig:qualitative}
\end{figure*}

\begin{figure*}[t]
\centering
    \centering
    \includegraphics[width=\columnwidth]{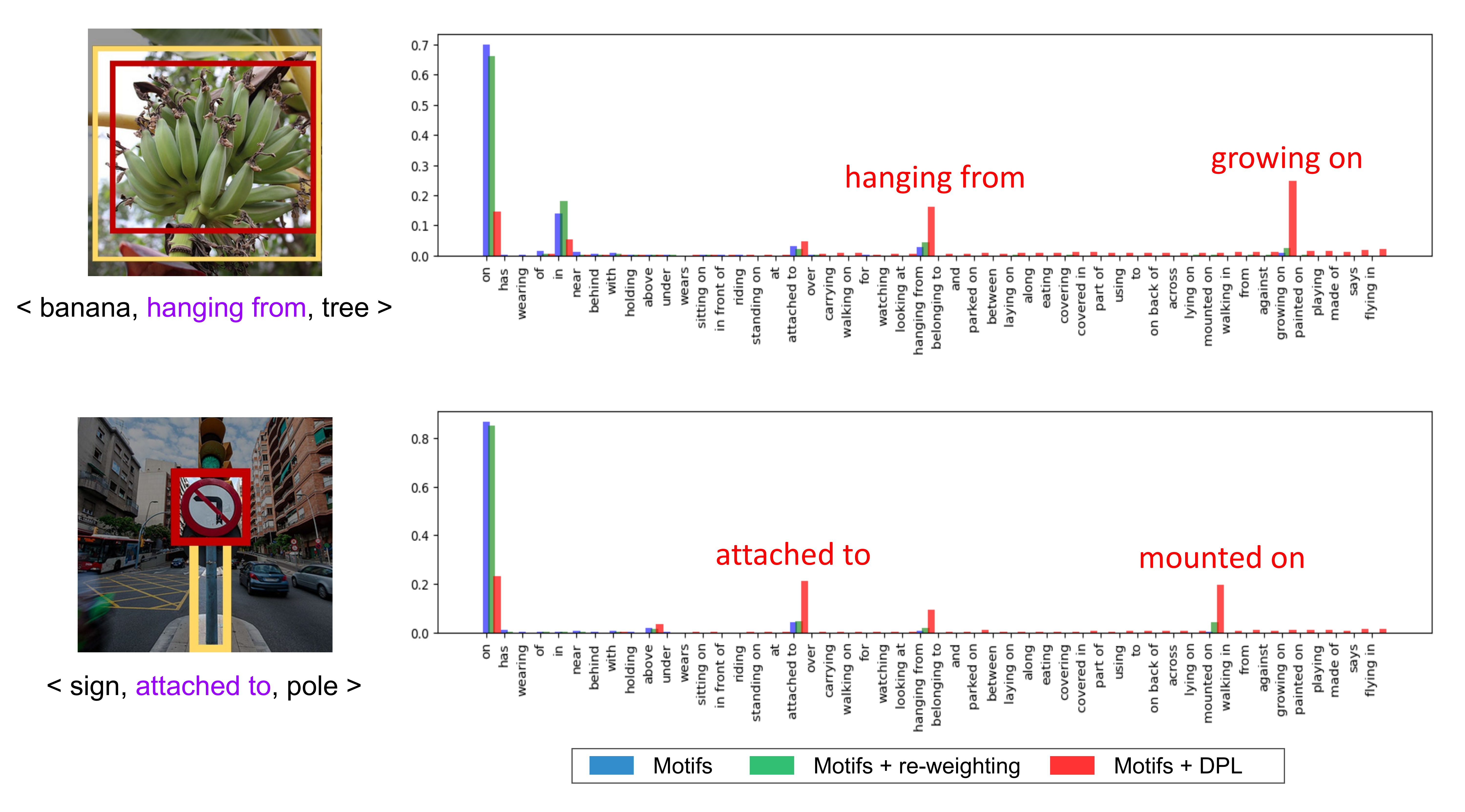}
    \captionof{figure}{ The predicted distribution over the predicate classes of Motifs, Motifs+re-weighting and Motifs+DPL. The purple-highlighted predicate below each image represents the ground truth predicate of the triplet.}
\label{fig:comp}
\end{figure*}

\subsubsection{Visualization of the semantic space.} 
In Fig.~\ref{fig:qualitative}, we visualize the representations of prototypes, relation features and samples.
We have the following observations:
\textbf{1)} In Fig.~\ref{fig:qualitative}(a), most of the relation features are indeed close to the prototype \texttt{on}, even though they are not labeled as  \texttt{on}. Therefore, when utilizing the conventional Euclidean distance-based prediction, most relationships tend to be classified with the predicate \texttt{on}, indicating biased predictions. 
\textbf{2)} Even among relation features labeled as \texttt{on} in Fig.~\ref{fig:qualitative}(a), their semantics are diverse. For example, the semantic of \texttt{on} in \texttt{<man, on, horse>} is relatively close to \texttt{riding}, while that in \texttt{<door, on, airplane>} is close to \texttt{attached to}. Moreover, the semantic of \texttt{on} in \texttt{<woman, on, sidewalk>} is close to \texttt{standing on} or \texttt{walking on}. 
\textbf{3)} In Fig.~\ref{fig:qualitative}(b), we observe that the samples of each prototype can indeed identify regions near \texttt{on} that are in fact associated with the semantics of other predicates. 
For example, the right side of the prototype \texttt{on} overlaps with \texttt{walking on}, while upper side of the prototype \texttt{on} overlaps with \texttt{riding}. This validates that DPL indeed captures the diverse semantics that a single predicate may exhibit.

\subsubsection{Comparisons with baselines.} 
Fig.~\ref{fig:comp} compares the actual predicted distribution of Motifs~\cite{zellers2018neural}, Motifs+reweight~\cite{chiou2021recovering}, and Motifs+DPL for two subject-object pairs.
We have the following observations:
\textbf{1) } Although all three models made an incorrect prediction on a test triplet \texttt{<banana, hanging from, tree>}, Motifs+DPL predicted \texttt{growing on} with the highest confidence followed by \texttt{hanging from}, while other two models predicted \texttt{on} with the highest confidence followed by \texttt{in}. We argue that our model not only provided a reasonable answer but also provides relatively unbiased predictions with fine-grained predicates.
\textbf{2) } For a test triplet \texttt{<sign, attached to, pole>}, Motifs+DPL effectively identifies suitable predicates to describe this relationship (e.g., \texttt{mounted on}, \texttt{attached to}), while other two models mostly focus on the predicate \texttt{on}, which again demonstrates that DPL is helpful for providing unbiased predictions.

\section{Conclusion}
In this paper, we tackle the problem of biased predictions caused by overlooking the semantic diversity of predicates. To this end, we proposed a model-agnostic Semantic Diversity-aware Prototype-based Learning (DPL) framework for unbiased SGG, aiming to identify the specific semantics of relationships that can be represented by the same predicate. Extensive experiments show that DPL not only significantly improves the performance of existing SGG models, but also yields reasonable qualitative results. 
There still remain limitations of DPL. Due to the small number of samples (i.e., $N$) used in this work, we fall short of learning precise regions in the semantic space. We expect further improvements in the performance when a large number of samples along with an appropriate $R$ value are utilized, which we leave as future work.

\section*{Acknowledgements}
This work was supported by the National Research Foundation of Korea(NRF) grant funded by the Korea government(MSIT) (RS-2024-00335098), Institute of Information \& communications Technology Planning \& Evaluation (IITP) grant funded by the Korea government(MSIT) (No.2022-0-00077), and National Research Foundation of Korea(NRF) funded by Ministry of Science and ICT (NRF-2022M3J6A1063021).

%
%
\bibliographystyle{splncs04}
\bibliography{main}

\title{Semantic Diversity-aware Prototype-based Learning for Unbiased Scene Graph Generation \\ (\textit{Supplementary})} 

\titlerunning{DPL}

\author{Jaehyeong Jeon\orcidlink{0009-0005-7622-6991} \and
Kibum Kim\orcidlink{0000-0002-7381-019X} \and
Kanghoon Yoon\orcidlink{0000-0001-6947-2944} \and
Chanyoung Park\thanks{Corresponding author.}\orcidlink{0000-0002-5957-5816}}

\authorrunning{J. Jeon et al.}

\institute{Korea Advanced Institute of Science and Technology (KAIST), South Korea \email{\{wogud405, kb.kim, ykhoon08, cy.park\}@kaist.ac.kr}}

\appendix
\newpage
\maketitle

\section{Additional Experiments}

\subsection{No-Graph Constraint Mean Recall}
Even within a single subject-object pair, there can exist multiple predicates representing their relationship. Therefore, instead of exclusively focusing on the predicate with the highest score for each pair, which is reflected in mR@K, it might be more appropriate to utilize a no-graph constraint mean recall (ng-mR@K) that considers all predicates. Table~\ref{tab:ngVG_table} shows that DPL exhibits significantly better performance compared to the existing the state-of-the-art models.

\begin{table}[h]
\caption{Performance (\%) comparisons in ng-mR@K on the VG dataset. The ng-mR@K performances of the existing models are reported by~\cite{chiou2021recovering, jung2023devil}.}
\centering
\resizebox{0.9\textwidth}{!}{
    \setlength{\tabcolsep}{4pt}
    \renewcommand{\arraystretch}{1.20}
        \begin{tabular}{l|ccc|ccc|ccc}
            \toprule
            \multicolumn{1}{c|}{\multirow{2}{*}{Model}} & \multicolumn{3}{c|}{PredCls}                                & \multicolumn{3}{c|}{SGCls}                               & \multicolumn{3}{c}{SGDet}                                \\ \cmidrule{2-10} 
            \multicolumn{1}{c|}{}                       & \multicolumn{1}{l}{ng-mR@20} & \multicolumn{1}{l}{ng-mR@50} & \multicolumn{1}{l|}{ng-mR@100} & \multicolumn{1}{l}{ng-mR@20} & \multicolumn{1}{l}{ng-mR@50} & \multicolumn{1}{l|}{ng-mR@100} & \multicolumn{1}{l}{ng-mR@20} & \multicolumn{1}{l}{ng-mR@50} & \multicolumn{1}{l}{ng-mR@100} \\ 
            \midrule
            Motifs~\cite{zellers2018neural}                                      
            & 19.9                  & 32.8                   & 44.7                   & 11.3                  & 19.0                     & 25.0                   & 7.5                  & 12.5                     & 16.9                  \\
            + Rwt~\cite{chiou2021recovering}                                      & 20.5                  & 33.5                   & 44.4                   & 12.6                  & 19.1                     & 24.3                   & 8.0                  & 12.9                     & 16.8                  \\
            
            + TDE~\cite{tang2020unbiased}                                       & 18.7                  & 29.0                   & 38.2                   & 10.7                  & 16.1                   & 21.1                   & 7.4                  & 11.2                     & 14.9                  \\
            + PCPL~\cite{yan2020pcpl}                                      
            & 25.6                  & 38.5                   & 49.3                   & 13.1                  & 19.9                   & 25.6                   & 9.8                  & 14.8                   & 19.6                  \\
            \rowcolor{gainsboro} + \textbf{\proposed}                   & \textbf{31.3}   & \textbf{46.8}   & \textbf{59.8}        & \textbf{18.5}   & \textbf{26.2}   & \textbf{32.0}        & \textbf{10.1}   & \textbf{15.3}   & \textbf{20.0}        \\ \midrule
            VCTree~\cite{tang2019learning}                            
            & 21.4                  & 35.6                   & 47.8                   & 14.3                  & 23.3                     & 31.4                  & 7.5                  & 12.5                     & 16.7                  \\
            + Rwt~\cite{chiou2021recovering}                           
            & 20.6                  & 32.5                   & 41.6                   & 14.1                  & 21.3                     & 27.8                  & 8.0                  & 12.1                     & 15.9                  \\
            + TDE~\cite{tang2020unbiased}                              
            & 20.9   & 32.4   & 41.5   & 12.4  & 19.1                  & 25.5   & 8.0    & 12.9   & 16.8                  \\
            + PCPL~\cite{yan2020pcpl}                                
            & 25.1   & 38.5   & 49.3   & 17.2  & 25.9                  & 32.7   & 9.9    & 15.1   & 19.9                  \\
            \rowcolor{gainsboro} + \textbf{\proposed}                   & \textbf{32.3}  & \textbf{48.7}  & \textbf{61.7}         & \textbf{19.6}  & \textbf{27.9}  & \textbf{33.9}          & \textbf{10.6}  & \textbf{16.3}  & \textbf{21.1}          \\  \bottomrule
        \end{tabular}
}
\label{tab:ngVG_table}
\end{table}

\subsection{Ablation Studies on GQA dataset}
We conduct the ablation studies of hyperparameters on GQA dataset, i.e., the number of samples $N$ and $R$ in Equation 6. Firstly, Table.~\ref{tab:Ng_table} presents the results for the hyperparameter $N$. Similar to the results on the VG dataset, the best performance is achieved when $N$=20. Secondly, as shown in Table.~\ref{tab:Rg_table}, we observe that when $R$ is too large or too small, the model fails to learn the appropriate range for each predicate, leading to a decrease in performance. For $R$, similar to the VG dataset, the best performance is observed when $R$=1.0.
\begin{table}[t]
    \centering
    \begin{minipage}[t]{0.5\textwidth}
        \centering
        \setlength{\tabcolsep}{5pt} 
        \renewcommand{\arraystretch}{1.0}
        \caption{Ablation studies on $N$.}
        \resizebox{0.85\textwidth}{!}{
            \begin{tabular}{c|ccc}
                \toprule
                \multirow{2}{*}{$N$} & \multicolumn{3}{c}{PredCls}                                                                                 \\ \cline{2-4} 
                                   & \multicolumn{1}{l}{R@50/100}    & \multicolumn{1}{l|}{mR@50/100}            & \multicolumn{1}{l}{F@50/100} \\ \midrule
                1                  & \multicolumn{1}{c|}{43.3 / 46.8} & \multicolumn{1}{c|}{30.7 / 33.2} & 35.9 / 38.8                   \\
                5                  & \multicolumn{1}{c|}{50.0 / 51.9} & \multicolumn{1}{c|}{30.9 / 33.2}          & 38.2 / 40.5                   \\
                10                 & \multicolumn{1}{c|}{52.2 / 53.7} & \multicolumn{1}{c|}{30.7 / 33.1}          & 38.7 / 41.0                   \\
                20                 & \multicolumn{1}{c|}{50.3 / 52.3} & \multicolumn{1}{c|}{\textbf{31.6} / \textbf{33.9}}          & \textbf{38.8} / \textbf{41.1}          \\
                40                 & \multicolumn{1}{c|}{52.0 / 53.7} & \multicolumn{1}{c|}{30.7 / 33.0}          & 38.6 / 40.9                   \\ 
                60                 & \multicolumn{1}{c|}{55.3 / 57.0} & \multicolumn{1}{c|}{29.5 / 31.6}          & 38.5 / 40.7          \\              
                \bottomrule
            \end{tabular}
        }
        \label{tab:Ng_table}
    \end{minipage}%
    \hfill
    \begin{minipage}[t]{0.5\textwidth}
        \centering
        \setlength{\tabcolsep}{5pt} 
        \renewcommand{\arraystretch}{1.15}
        \caption{Ablation studies on $R$.}
        \resizebox{0.87\textwidth}{!}{
            \begin{tabular}{c|ccc}
            \toprule
            \multirow{2}{*}{$R$} & \multicolumn{3}{c}{PredCls}                                                                        \\ \cline{2-4} 
                               & \multicolumn{1}{l|}{R@50/100}    & \multicolumn{1}{l|}{mR@50/100}   & \multicolumn{1}{l}{F@50/100} \\ \midrule
            0.6                & \multicolumn{1}{c|}{40.8 / 46.9} & \multicolumn{1}{c|}{27.2 / 30.2} & 32.6 / 36.7                   \\
            0.8                & \multicolumn{1}{c|}{49.5 / 54.1} & \multicolumn{1}{c|}{27.1 / 30.0} & 35.0 / 38.6                   \\
            1.0                & \multicolumn{1}{c|}{50.3 / 52.3} & \multicolumn{1}{c|}{\textbf{31.6 / 33.9}} & \textbf{38.8 / 41.1}          \\
            1.2                & \multicolumn{1}{c|}{61.5 / 63.0} & \multicolumn{1}{c|}{24.2 / 25.9} & 34.7 / 36.7                   \\
            1.4                & \multicolumn{1}{c|}{63.5 / 65.5} & \multicolumn{1}{c|}{20.9 / 22.9} & 31.4 / 33.9                   \\ \bottomrule
            \end{tabular}
        }
        \label{tab:Rg_table}
    \end{minipage}
\end{table}

\subsection{Assigning $N$ Based on Prior Knowledge of Frequencies}
We conduct experiments using different numbers of $N$ for each predicate in Table 3 of the main paper. However, these experiments involve assigning arbitrarily high or low values of $N$ to the head and tail classes. Therefore, we further explore this in Table~\ref{tab:Ndf_table} by assigning different values of $N$ to each predicate based on their frequencies. We assigned $N$ based on log-transformed frequency, and compared cases with an average $N$=20 and $N$=2. The case (avg $N$=20) is to allow a fair comparison with the case where each predicate is given exactly 20 samples. The reason for comparing the case (avg $N$=2) is to check whether comparable performance can be achieved with fewer samples. Specifically, in that case (avg $N$=2), \texttt{on} has $N$=6 while most tail classes are given $N$=1. In Table~\ref{tab:Ndf_table}, we can confirm that while R@K and mR@K differ slightly, the overall performance (F@K and M@K) is nearly identical between the case where N=20 for all predicates and the case where N is assigned based on frequency with an average of $N$=20. Furthermore, even with an average $N$=2, the performance does not show a significant decrease compared to other cases, demonstrating that DPL can be used more efficiently.

\begin{table}[h]
    \centering
    \caption{Ablation studies of different $N$ based on frequency.}
    \setlength{\tabcolsep}{5pt} 
    \renewcommand{\arraystretch}{1.2}
    \resizebox{0.85\columnwidth}{!}{
        \begin{tabular}{l|cccc}
            \toprule
            \multirow{2}{*}{Model} & \multicolumn{4}{c}{PredCls} \\ \cline{2-5} 
                               & \multicolumn{1}{l|}{R@50/100}    & \multicolumn{1}{l|}{mR@50/100}            & \multicolumn{1}{l|}{F@50/100} & M@50/100 \\ \midrule
            Motifs~[\textcolor{blue}{1}]                  & \multicolumn{1}{c|}{\textbf{65.2} / \textbf{67.0}} & \multicolumn{1}{c|}{14.8 / 16.1} & \multicolumn{1}{c|}{24.1 / 26.0} & 40.0 / 41.6   \\
            + DPL($N=20$ for all predicates)                  & \multicolumn{1}{c|}{54.4 / 56.3} & \multicolumn{1}{c|}{\textbf{33.7} / \textbf{37.4}}          & \multicolumn{1}{c|}{\textbf{41.6} / \textbf{44.9}} & 44.1 / 46.9                  \\ 
            + DPL($N$ is set based on freq. (avg. $N=20$))     & \multicolumn{1}{c|}{56.7 / 58.7} & \multicolumn{1}{c|}{32.5 / 35.9} & \multicolumn{1}{c|}{41.3 / 44.6} & \textbf{44.6} / \textbf{47.3} \\
            + DPL($N$ is set based on freq. (avg. $N=2$))     & \multicolumn{1}{c|}{53.3 / 55.1} & \multicolumn{1}{c|}{33.2 / 36.7} & \multicolumn{1}{c|}{40.9 / 44.1} & 43.3 / 45.9
            \\ \bottomrule        
        \end{tabular}
    }
    \label{tab:Ndf_table}
\end{table}

\subsection{Training Time Analysis}
In this paper, we chose $N$ to be a number within 100. The reason behind this decision is the strong correlation between $N$ and computational complexity. Table.~\ref{tab:time_table} presents the training time while changing $N$ under the same GPU environment (NVIDIA A6000 48GB). When $N$ is less than 100, the difference in time compared to $N$=1 is not substantial. However, if $N$ exceeds 100, the computational cost significantly increases, leading to longer training times. Therefore, even though the performance is expected to improve with a large $N$ given an appropriate $R$, it may not be favorable when considering the cost-effectiveness of DPL.

\begin{table}[h]
    \begin{minipage}[t]{0.33\textwidth}
        \centering
        \setlength{\tabcolsep}{5pt} 
        \renewcommand{\arraystretch}{1.15}
        \caption{Training time comparision of DPL on PredCls.}
        \resizebox{1.0\textwidth}{!}{
            \begin{tabular}{lc}
                \toprule
                \multicolumn{1}{c}{Model} & Time / img (sec)        \\ \midrule 
                \multicolumn{1}{l|}{Motifs~\cite{zellers2018neural}}  & 0.115 \\
                \multicolumn{1}{l|}{+ \proposed (N=1)}  & 0.158 \\
                \multicolumn{1}{l|}{+ \proposed (N=10)}  & 0.159 \\
                \multicolumn{1}{l|}{+ \proposed (N=100)}  & 0.162 \\
                \multicolumn{1}{l|}{+ \proposed (N=1000)}  & 0.195 \\
                \multicolumn{1}{l|}{+ \proposed (N=10000)}  & 0.472 \\
                \bottomrule     
            \end{tabular}
        }
        \label{tab:time_table}
    \end{minipage}
    \hfill
    \begin{minipage}[t]{0.65\textwidth}
        \centering
        \captionof{figure}{Recall@100 comparisons for all predicate classes on the PredCls task.}
        \resizebox{1.0\textwidth}{!}{
            \centering
            \includegraphics[width=0.95\columnwidth]{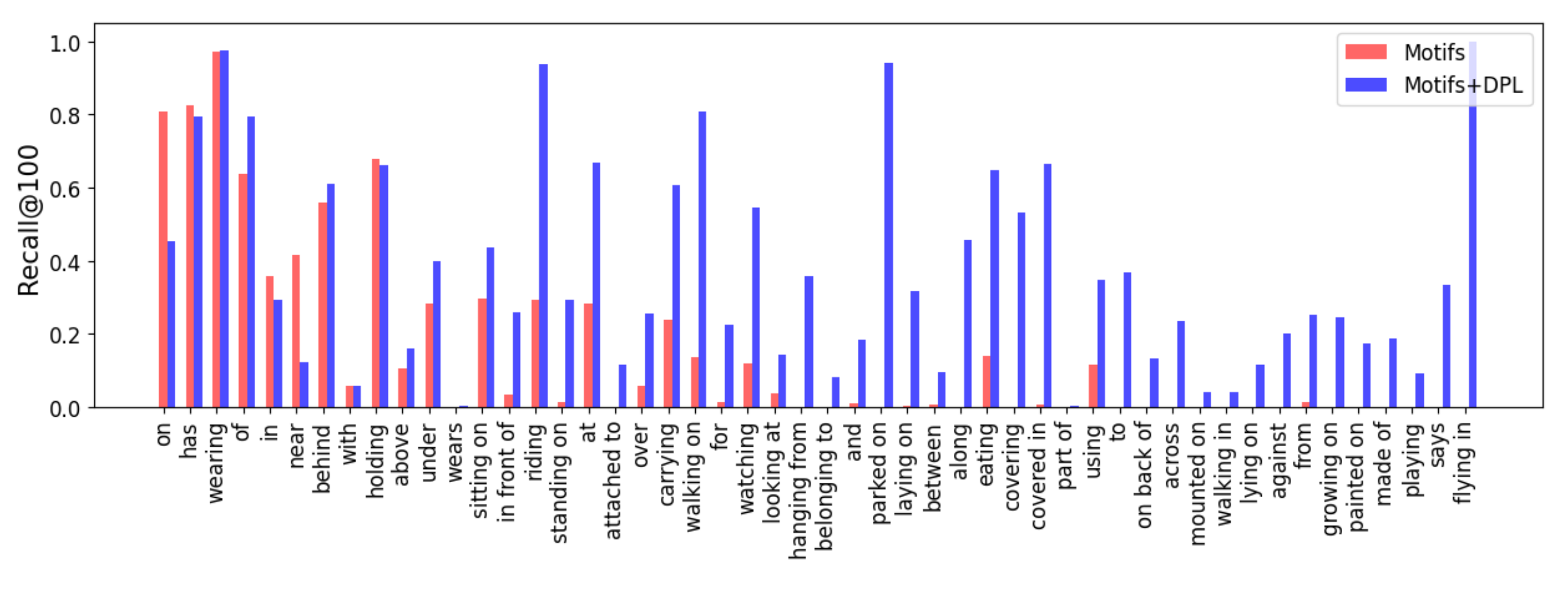}
        }
        \label{fig:recall}
    \end{minipage}%

\end{table}

\section{Additional Qualitative Results}
\subsection{Visualization of Recall for Each Predicate}
For a detailed analysis of performance across different predicate class, we present the recall of Motifs and Motifs+DPL for each predicate in Fig.~\ref{fig:recall}. These results are obtained from the PredCls task on the VG dataset. DPL generally exhibits significantly better performance than Motifs~\cite{zellers2018neural} across most predicates, except for a few head classes (e.g., \texttt{on}, \texttt{near}). In particular, we observe a notable improvement in the performance of predicates related to \texttt{on}, such as \texttt{riding}, \texttt{walking on}, \texttt{sitting on}, and \texttt{parked on}. This suggests that understanding the semantic diversity of \texttt{on} helps in estimating more informative predicates. Similarly, there is a significant improvement in the predicates related to \texttt{near}, such as \texttt{in front of} and \texttt{at}. This indicates that the model effectively captures the semantic diversity of \texttt{near}.

\subsection{Probability distributions Comparison}

In Fig.~\ref{fig:comp2}, we provide additional predicted probability distributions of relationships. DPL outperforms other models in identifying more appropriate predicates for representing relationships. Even if the predicted predicates differ from the ground truth, they still predict reasonable predicates.

\begin{figure}[t]
\centering
    \centering
    \includegraphics[width=\columnwidth]{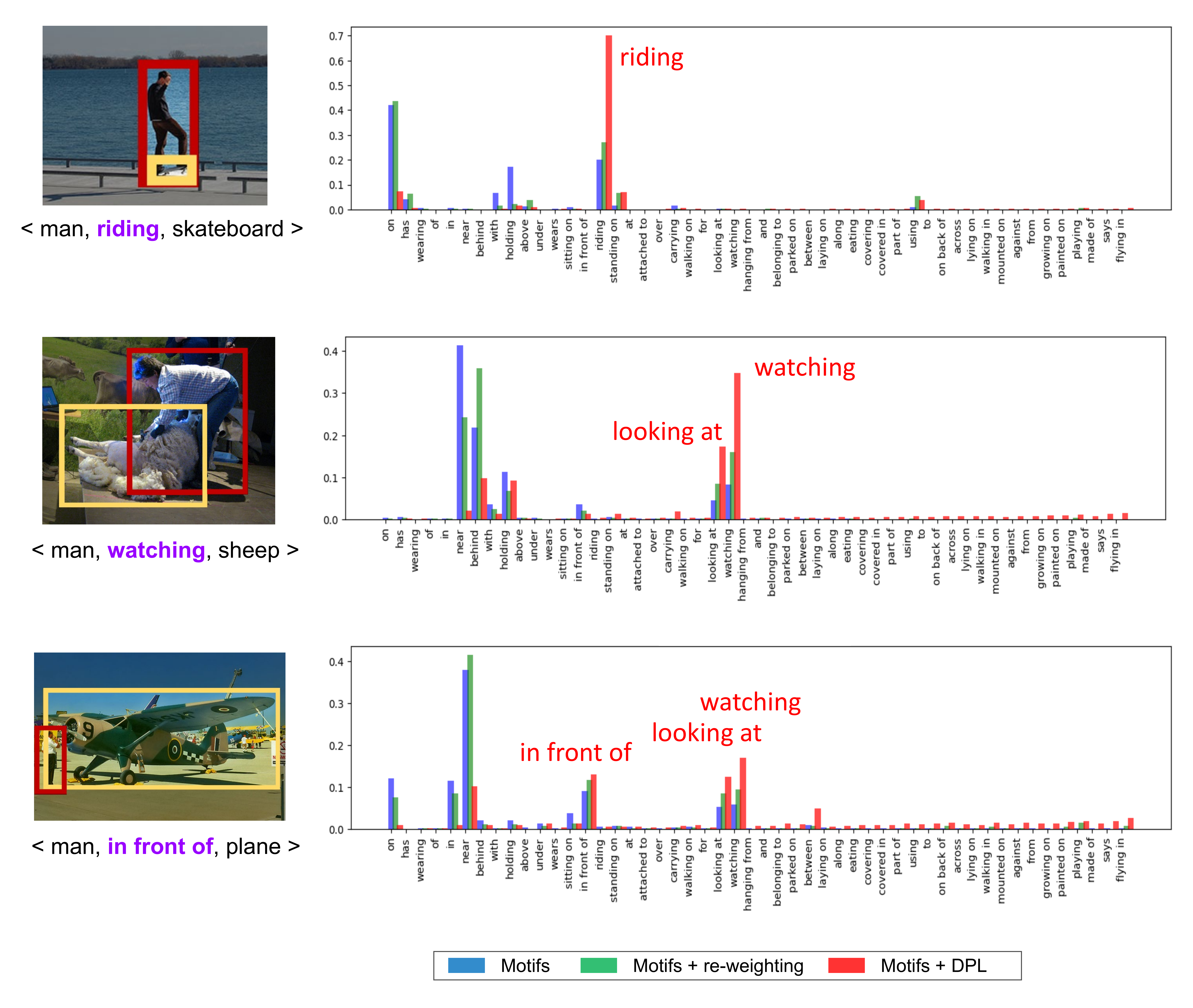}
    \captionof{figure}{The predicted distribution over the predicate classes of Motifs, Motifs+re-weighting and Motifs+DPL. The purple-highlighted predicate below each image represents the ground truth predicate of the triplet.}
\label{fig:comp2}
\end{figure}

\section{Potential Negative Impact and Limitations}
Although our study employs prototypes, the core concept lies in understanding semantic diversity through probabilistic sampling. However, there might be a misconception that the unbiased SGG achieved in this paper was solely attributed to the prototypes. As shown in the ablation study on each component, it is evident that prototypes alone are insufficient for unbiased predictions. Additionally, one limitation of our study is its dependence on the performance of the object detector, which consequently leads to relatively lower performance in the SGDet task.

\end{document}